\documentclass{article} 
\usepackage{iclr2015,times}
\usepackage{hyperref}
\usepackage{url}
\usepackage{graphicx}
\usepackage{authblk}
\usepackage{subcaption}
\usepackage{amsmath}
\usepackage{xcolor}

\usepackage{tikz}
\usetikzlibrary{calc}
\usetikzlibrary{external}
\usepackage{pgfplots}
\pgfplotsset{compat=newest}
\pgfplotsset{plot coordinates/math parser=false}
\pgfplotsset{
compat=newest,
/pgfplots/myylabel absolute/.style={%
  /pgfplots/every axis y label/.style={at={(0,0.5)},xshift=#1,rotate=90},
  /pgfplots/every y tick scale label/.style={
    at={(0,1)},above right,inner sep=0pt,yshift=0.3em
   }
  }
}
\pgfplotsset{
  log x ticks with fixed point/.style={
      xticklabel={
        \pgfkeys{/pgf/fpu=true}
        \pgfmathparse{exp(\tick)}%
        \pgfmathprintnumber[fixed relative, precision=3]{\pgfmathresult}
        \pgfkeys{/pgf/fpu=false}
      }
  },
  log y ticks with fixed point/.style={
      yticklabel={
        \pgfkeys{/pgf/fpu=true}
        \pgfmathparse{exp(\tick)}%
        \pgfmathprintnumber[fixed relative, precision=3]{\pgfmathresult}
        \pgfkeys{/pgf/fpu=false}
      }
  }
}

\newcommand{\fig}[1]{Figure~\ref{fig:#1}}
\newcommand{\sect}[1]{Section~\ref{sect:#1}}
\newcommand{\tab}[1]{Table~\ref{tab:#1}}
\newcommand{\eq}[1]{(\ref{eq:#1})}

\newcommand{\todo}[1][]{\@latex@warning{TODO #1}\fbox{TODO\dots}}

\title{Speeding-up Convolutional Neural Networks\\Using Fine-tuned CP-Decomposition}

\author[1,2]{Vadim Lebedev}
\author[1]{Yaroslav Ganin}
\author[1,3]{Maksim Rakhuba}
\author[1,4]{Ivan Oseledets}
\author[1]{Victor Lempitsky}
\affil[1]{Skolkovo Institute of Science and Technology (Skoltech), Moscow, Russia}
\affil[2]{Yandex, Moscow, Russia}
\affil[3]{Moscow Institute of Physics and Technology, Moscow Region, Russia}
\affil[4]{Institute of Numerical Mathematics RAS, Moscow, Russia}
\newlength\figureheight
\newlength\figurewidth

\iclrconference 

\begin{document}

\maketitle

\begin{abstract}
We propose a simple two-step approach for speeding up convolution layers within large convolutional neural networks based on tensor decomposition and discriminative fine-tuning. Given a layer, we use non-linear least squares to compute a low-rank CP-decomposition of the 4D convolution kernel tensor into a sum of a small number of rank-one tensors. At the second step, this decomposition is used to replace the original convolutional layer with a sequence of four convolutional layers with small kernels. After such replacement, the entire network is fine-tuned on the training data using standard backpropagation process.

We evaluate this approach on two CNNs and show that it is competitive with previous approaches, leading to higher obtained CPU speedups at the cost of lower accuracy drops for the smaller of the two networks. Thus, for the 36-class character classification CNN, our approach obtains a 8.5x CPU speedup of the whole network with only minor accuracy drop ($1\%$ from $91\%$ to $90\%$). For the standard ImageNet architecture (AlexNet), the approach speeds up the second convolution layer by a factor of 4x at the cost of $1\%$ increase of the overall top-5 classification error.
\end{abstract}

\section{Introduction}
\label{sect:intro}

Over the course of two years, Convolutional neural networks (CNNs) \citep{LeCun89} have revolutionized computer vision and became ubiquituous through a range of computer vision applications. In many ways, this breakthrough has become possible through the acceptance of new computational tools, most notably GPUs \citep{Krizhevsky12}, but also CPU clusters \citep{Dean12} and FPGAs \citep{Farabet11}. On the other hand, there is a rising interest to deploying CNNs on low-end architectures, such as standalone desktop/laptop CPUs, mobile processors, and CPU in robotics. On such processors, the computational cost of applying, yet alone training a CNN might pose a problem, especially when real-time operation is required.

The key layer of CNNs that distinguishes them from other neural networks and enables their recent success is the convolution operation. Convolutions dominate the computation cost associated with training or testing a CNN (especially for newer architectures such as \citep{Simonyan14}, which tend to add more and more convolutional layers often at the expense of  fully-connected layers). Consequently, there is a strong interest to the task of improving the efficiency of this operation \citep{Chintala14, Mathieu13, Chellapilla06}. An efficient implementation of the convolution operation is one of the most important elements of all major CNN packages.

A group of recent works have achieved significant speed-ups of CNN convolutions by applying tensor decompositions. In more detail, recall that a typical convolution in a CNN takes as an input a 3D tensor (array) with the dimensions corresponding to the two spatial dimensions, and to different image maps. The output of a convolution is another similarly-structured 3D tensor. The convolution kernel itself constitutes a 4D tensor with dimensions corresponding to the two spatial dimensions, the input image maps, and the output image maps. Exploiting this tensor structure, previous works \citep{Denton14,Jaderberg14} have suggested different tensor decomposition schemes to speed-up convolutions within CNNs. These schemes are applied to the kernel tensor and generalize previous 2D filter approximations in computer vision like \citep{Rigamonti13}.

In this work, we further investigate tensor decomposition in the context of speeding up convolutions within CNNs. We use a standard (in tensor algebra) low-rank CP-decomposition (also known as PARAFAC or CANDECOMP) and an efficient optimization approach (non-linear least squares) to decompose the full kernel tensor. 

We demonstrate that the use of the CP-decomposition on a full kernel tensor has the following important advantages:

\begin{itemize}
\item {\bf Ease of the decomposition implementation.} CP-decomposition is one of the standard tools in tensor linear algebra with well understood properties and mature implementations. Consequently, one can use existing efficient algorithms such as NLS to compute it efficiently.

\item {\bf Ease of the CNN implementation.} CP-decomposition approximates the convolution with a 4D kernel tensor by the sequence of four convolutions with small 2D kernel tensors. All these lower dimensional convolutions represent standard CNN layers, and are therefore easy to insert into a CNN using existing CNN packages (no custom layer implementation is needed).

\item {\bf Ease of fine-tuning.} Related to the previous item, once a convolutional layer is approximated and replaced with a sequence of four convolutional layers with smaller kernels, it is straight-forward to fine-tune the \emph{entire} network on training data using back-propagation.

\item {\bf Efficiency.} Perhaps most importantly, we found that the simple combination of the full kernel CP-decomposition and global fine-tuning can result in better speed-accuracy trade-off for the approximated networks than the previous methods.
\end{itemize}

The CNNs obtained in our method are somewhat surprisingly accurate, given that the number of parameters is reduced several times compared to the initial networks. Practically, this reduction in the number of parameters means more compact networks with reduced memory footprint, which can be important for architectures with limited RAM or storage memory. Such memory savings can be especially valuable for feed-forward networks that include convolutions with spatially-varying kernels (``locally-connected'' layers).

On the theoretical side, these results confirm the intuition that modern CNNs are over-parameterized, i.e.\ that the sheer number of parameters in the modern CNNs are not needed to store the information about the classification task but, rather, serve to facilitate convergence to good local minima of the loss function.

Below, we first discuss previous works of \citep{Rigamonti13,Jaderberg14,Denton14} and outline the differences between them and our approach in \sect{related}. We then review the CP-decomposition and give the details of our approach in \sect{method}. The experiments on two CNNs, namely the character classification CNN from \citep{Jaderberg14} and \citep{Jaderberg14a}  and AlexNet from the \texttt{Caffe} package \citep{caffe}, follow in \sect{experiments}. We conclude with the summary and the discussion in \sect{discussion}.
\section{Related work}
\label{sect:related}

\begin{figure}[t]
\centering
\hspace{-4mm}\begin{tabular}{cc}
\includegraphics[height=3cm]{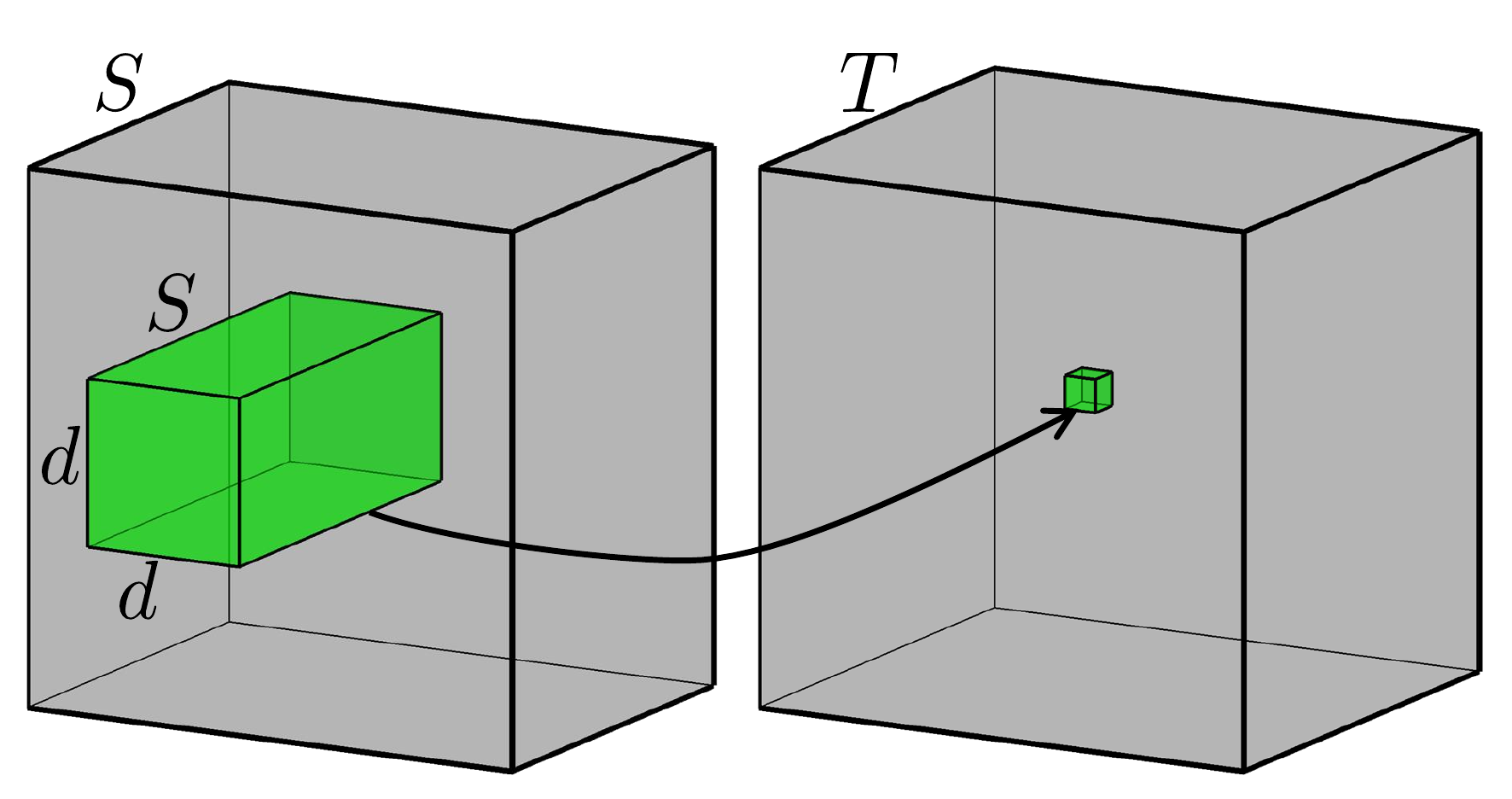}&
\hspace{-4mm}\includegraphics[height=3cm]{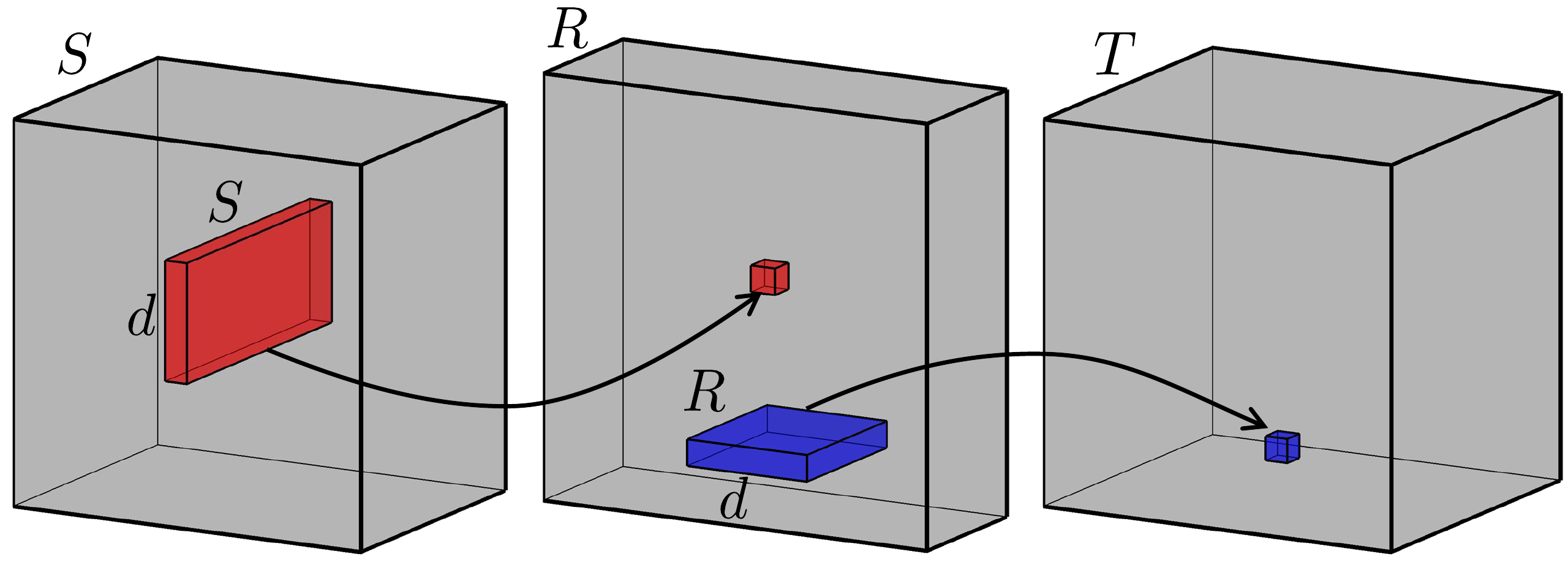}\\
(a) Full convolution&\hspace{-4mm}(b) Two-component decomposition \citep{Jaderberg14}\\
\end{tabular}
\begin{tabular}{c}
\includegraphics[height=3cm]{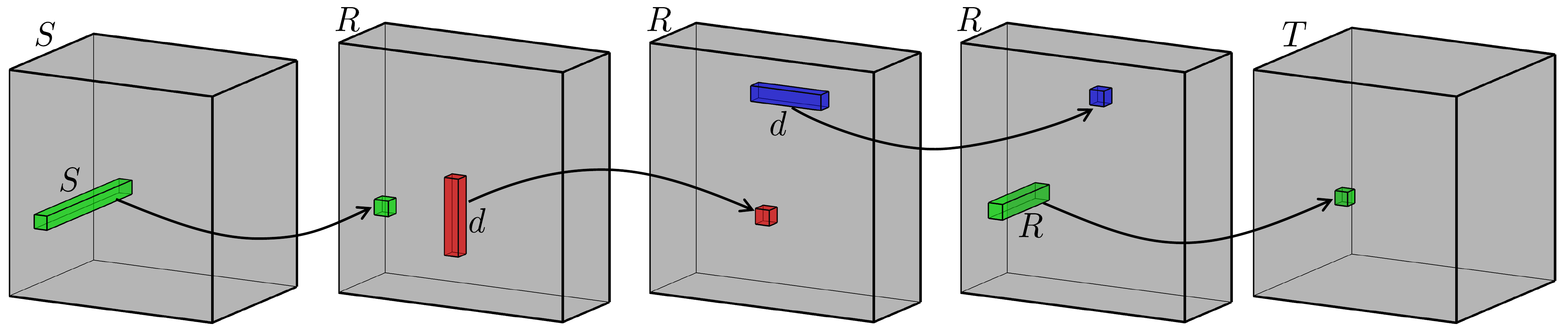}\\
(c) CP-decomposition
\end{tabular}
\caption{ {\bf Tensor decompositions for speeding up a generalized convolution.} Gray boxes correspond to 3D tensors (map stacks) within a CNN, with the two frontal sides corresponding to spatial dimensions. Arrows show linear mappings and demonstrate how scalar values on the right (small boxes corresponding to single elements of the target tensor) are computed. Initial \textbf{full convolution} (a) computes each element of the target tensor as a linear combination of the elements of a 3D subtensor that spans a spatial $d\times d$ window over all input maps.  {\bf \citet{Jaderberg14}} (b) approximate the initial convolution as a composition of two linear mappings with the intermediate map stack having $R$ maps (where $R$ is the rank of the decomposition). Each of the two mappings computes each target value based on a spatial window of size $1\times{}d$ or $d\times{}1$ in all input maps. Finally, \textbf{CP-decomposition} (c) used in our approach approximates the convolution as a composition of four convolutions with small kernels, so that a target value is computed based on a 1D-array that spans either one pixel in all input maps, or a 1D spatial window in one input map. }
\label{fig:decomp}
\end{figure}

{\bf Decompositions for convolutions.} Using low-rank decomposition to accelerate convolution was suggested by \citet{Rigamonti13} in the context of codebook learning. In \citep{Rigamonti13}, a bank of 2D or 3D filters $X$ is decomposed into linear combinations of a shared bank of separable (decomposable) filters $Y$. The decompositions of filters within $Y$ are independent (components are not shared).

\citet{Jaderberg14} evaluated the decomposition \citep{Rigamonti13} in the context of CNNs and furthermore suggested a more efficient  decomposition (\fig{decomp}b) that effectively approximates the 4D kernel tensor as a composition (product) of two 3D tensors. In the experiments, \citet{Jaderberg14} have demonstrated the advantage of this scheme over \citep{Rigamonti13}. In a sequel, when refering to \citep{Jaderberg14} we imply this two-component decomposition.

Once the decomposition is computed, \citet{Jaderberg14} perform ``local'' fine-tuning that minimizes the deviation between the full and the approximated convolutions outputs on the training data. Differently from \citet{Jaderberg14}, our method fine-tunes the entire network based on the original discriminative criterion.
While \citep{Jaderberg14} reported that such discriminative fine-tuning was inefficient for their scheme, we found that in our case it works well, even when CP-decomposition has large approximation error. Below, we provide a theoretical complexity comparison and empirical comparison of our scheme with \citep{Jaderberg14}.

In the work that is most related to ours, \citet{Denton14} have suggested a scheme based on the CP-decomposition of parts of the kernel tensor obtained by \textit{biclustering} (alongside with a different decompositions for the first convolutional layer and the fully-connected layers). Biclustering of \citep{Denton14} splits the two non-spatial dimensions into subgroups, and reduces the effective ranks in the CP-decomposition. CP-decompositions of the kernel tensor parts in \citep{Denton14} have been computed with the greedy approach\footnote{Note that the alternating least squares process mentioned in \citep{Denton14} refers to computing the best next rank-1 tensor, but the outer process of adding rank-1 tensors is still greedy.}. 
Our approach essentially simplifies that of \citet{Denton14} in that we do not perform biclustering and apply CP-decomposition directly to the full convolution kernel tensor. On the other hand, we replace greedy computation of CP-decomposition with non-linear least squares. Finally, as discussed above, we fine-tune the complete network by backpropagation, whereas \cite{Denton14} only fine-tunes the layers above the approximated one.


\section{Method}
\label{sect:method}

Overall our method is a conceptually simple two-step approach: (1) take a convolutional layer and decompose its kernel using CP-decomposition, (2) fine-tune the entire network using backpropagation. If necessary, proceed to another layer. Below, we review the CP-decomposition, which is at the core of our method, and provide the details for the two steps of our approach.

\subsection{CP-decomposition review}

Tensor decompositions are a natural way to generalize low-rank approach to multidimensional case.
Recall that a low-rank decomposition of a matrix $A$ of size $n \times m$ with rank $R$ is given by: 
\begin{equation}
    A(i, j) = \sum_{r = 1}^R A_1(i, r) A_2(j, r), \quad i = \overline{1,n}, \quad j = \overline{1,m},
\end{equation}
and leads to the idea of separation of variables.
The most straightforward way to separate variables in case of many dimensions is to use the canonical polyadic decomposition (CP-decomposition, also called as CANDECOMP/PARAFAC model) \citep{kolda-review-2009}.
For a $d$-dimensional array $A$ of size $n_1 \times \dots \times n_d$ a CP-decomposition has the following form
\begin{equation}
A(i_1, \dots, i_d) = 
    \sum_{r = 1}^R A_1(i_1, r) \dots A_d(i_d, r),\label{eq:dcmp}
\end{equation}
where the minimal possible $R$ is called canonical rank.
The profit of this decomposition is that we need to store only 
$(n_1 + \dots + n_d) R$ elements instead of the whole tensor with $n_1 ... n_d$ elements.

In two dimensions, the low-rank approximation can be computed in a stable way by using singular value decomposition (SVD) or, if the matrix is large, by rank-revealing algorithms.
Unfortunately, this is not the case for the CP-decomposition when $d>2$, as there is no finite algorithm for determining canonical rank of a tensor \citep{kolda-review-2009}. 
Therefore, most algorithms approximate a tensor with different values of $R$ until the approximation error becomes small enough.
This leads to the point that for finding good low-rank CP-decomposition certain tricks have to be used.
A detailed survey of methods for computing low-rank CP-decompositions can be found in \citep{tomasi-comparison-2006}.
As a software package to calculate the CP-decomposition we used \texttt{Tensorlab} \citep{sorber-tensorlab-2014}.
For our purposes, we chose non-linear least squares (NLS) method, which minimizes the L2-norm of the approximation residual (for a user-defined fixed $R$) using Gauss-Newton optimization.

Such NLS optimization is capable of obtaining much better approximations than the strategy of greedily finding best rank-1 approximation of the residual vectors used in \cite{Denton14}. The fact that the greedy rank-1 algorithm may increase tensor rank can be found in \citep{Stegeman10, Kof02}.
We also give a simple example highlighting this advantage of the NLS in the Appendix.

\subsection{Kernel tensor approximation}

CNNs \citep{LeCun89} are feed-forward multi-layer architectures that map the input images to certain output vectors using a sequence of operations.
The units within CNN are organized as a sequence of 3D tensors (``map stacks'') with two spatial dimensions and the third dimension corresponding to different ``maps'' or ''channels''\footnote{These tensors are 4D if/when a CNN is applied to a batch of images, with the fourth dimension corresponding to different images in a batch. This extra dimension does not affect the derivation below and is therefore disregarded.}.

The most ``important'' and time-consuming operation within modern CNNs is the generalized convolution that maps an input tensor $U(\cdot,\cdot,\cdot)$ of size $X{\times}Y{\times}S$  into an output tensor $V(\cdot,\cdot,\cdot)$ of size $(X{-}d{+}1){\times}(Y{-}d{+}1){\times}T$ using the following linear mapping:
\begin{equation}
V(x,y,t) = \sum_{i=x-\delta}^{x+\delta} \; \sum_{j=y-\delta}^{y+\delta} \; \sum_{s=1}^S K(i-x+\delta,j-y+\delta,s,t)\, U(i,j,s) \label{eq:convfull}
\end{equation}

Here, $K(\cdot,\cdot,\cdot,\cdot)$ is a 4D kernel tensor of size $d{\times}d{\times}S{\times}T$ with the first two dimensions corresponding to the spatial dimensions, the third dimension corresponding to different input channels, the fourth dimension corresponding to different output channels. The spatial width and height of the kernel are denoted as $d$, while $\delta$ denotes ``half-width'' $(d-1)/2$ (for simplicity we assume square shaped kernels and even $d$). 

The rank-$R$ CP-decomposition \eq{dcmp} of the 4D kernel tensor has the form:
\begin{equation}
K(i,j,s,t) = \sum_{r=1}^R  K^x(i-x+\delta,r)\,  K^y(j-y+\delta,r)\, K^s(s,r)  \, K^t(t,r)\,,\label{eq:cp}
\end{equation}
where $K^x(\cdot,\cdot)$, $K^y(\cdot,\cdot)$, $K^s(\cdot,\cdot)$, $K^t(\cdot,\cdot)$ are the four components of the composition representing 2D tensors (matrices) of sizes $d{\times}R$, $d{\times}R$, $S{\times}R$, and $T{\times}R$ respectively.

Substituting \eq{cp} into \eq{convfull} and performing permutation and grouping of summands gives the following expression for the approximate evaluation of the convolution \eq{convfull}:
\begin{equation}
V(x,y,t) = \sum_{r=1}^{R} K^t(t,r) 
\left(\sum_{i=x-\delta}^{x+\delta} K^x(i-x+\delta,r)
\left(\sum_{j=y-\delta}^{y+\delta} K^y(j-y+\delta,r)
\left(\sum_{s=1}^S \, K^s(s,r) \, U(i,j,s)\right)\right)\right) \label{eq:appr}
\end{equation}

Based on \eq{appr}, the output tensor $V(\cdot,\cdot,\cdot)$ can be computed from the input tensor $U(\cdot,\cdot,\cdot)$ via a sequence of four convolutions with smaller kernels (\fig{decomp}c):
\begin{align}
U^s(i,j,r) &=\, \sum_{s=1}^S K^s(s,r)\, U(i,j,s) \label{eq:steps}\\
U^{sy}(i,y,r) &= \sum_{j=y-\delta}^{y+\delta} K^y(j-y+\delta,r) \, U^s(i,j,r) \label{eq:stepy}\\ 
U^{syx}(x,y,r) &= \sum_{i=x-\delta}^{x+\delta} K^x(i-x+\delta,r) \, U^{sy}(i,y,r) \label{eq:stepx}\\ 
V(x,y,t) &= \sum_{r=1}^{R} K^t(t,r)\,  U^{syx}(x,y,r)\,, \label{eq:stept}
\end{align}
where $U^s(\cdot,\cdot,\cdot)$, $U^{sy}(\cdot,\cdot,\cdot)$, and $U^{syx}(\cdot,\cdot,\cdot)$ are intermediate tensors (map stacks). 

\subsection{Implementation and Fine-tuning}

Computing $U^s(\cdot,\cdot,\cdot)$ from $U(\cdot,\cdot,\cdot)$ in \eq{steps} as well as $V(\cdot,\cdot,\cdot)$ from $U^{syx}(\cdot,\cdot,\cdot)$ in \eq{stept} represent so-called $1{\times}1$~convolutions (also used within ``network-in-network'' approach \citep{Lin13}) that essentially perform pixel-wise linear re-combination of input maps. Computing $U^{sy}(\cdot,\cdot,\cdot)$ from $U^s(\cdot,\cdot,\cdot)$ and  $U^{syx}(\cdot,\cdot,\cdot)$ from $U^{sy}(\cdot,\cdot,\cdot)$ in \eq{stepy} and \eq{stepx} are ``standard'' convolutions with small kernels that are ``flat'' in one of the two spatial dimensions.

We use the popular \texttt{Caffe} package \citep{caffe} to implement the resulting architecture, utilizing standard convolution layers for \eq{stepy} and \eq{stepx}, and an optimized $1{\times}1$~convolution layers for \eq{steps} and \eq{stept}. The resulting architecture is fine-tuned through standard backpropagation (with momentum) on training data. All network layers including layers above the approximated layer, layers below the approximated layer, and the four inserted convolutional layers participate in the fine-tuning. We found, however, that the gradients within the inserted layers are prone to gradient explosion, so one should either be careful to keep the learning rate low, or fix some or all of the inserted layers, while still fine-tuning layers above and below.

\subsection{Complexity analysis}

Initial convolution operation is defined by  $STd^2$ parameters (number of elements in the kernel tensor) and requires the same number of ``multiplication+addition'' operations per pixel.

For \citep{Jaderberg14} this number changes to $Rd(S+T)$, where $R$ is the rank of the decomposition (see \fig{decomp}b and \cite{Jaderberg14}). While the two numbers are not directly comparable, assuming that the required rank is comparable or several times smaller than $S$ and $T$ (e.g.\ taking $R \approx \frac{ST}{S+T}$), the scheme \citep{Jaderberg14} gives a reduction in the order of $d$ times compared to the initial convolution.

For \citep{Denton14} in the absence of bi-clustering as well as in the case of our approach, the complexity is $R(S+2d+T)$ (again, both for the number of parameters and for the number of ``multiplications+additions'' per output pixel). Almost always, $d \ll T$, which for the same rank gives a further factor of $d$ improvement in complexity over \citep{Jaderberg14} (and an order of $d^2$ improvement over the initial convolution when $R \approx \frac{ST}{S+T}$).

The bi-clustering in \citep{Denton14} makes a ``theoretical'' comparison with the complexity of our approach problematic, as on the one hand bi-clustering increases the number of tensors to be approximated, but on the other hand, reduces the required ranks considerably (so that assuming the same $R$ would not be reasonable). We therefore restrict ourselves to the empirical comparison.




\section{Experiments}
\label{sect:experiments}

In this section we test our approach on two network architectures, small character-classification CNN and a bigger net trained for ILSVRC. Most of our experiments are devoted to the approximation of single layers, when other layers remain intact apart from the fine-tuning.

We make several measurements to evaluate our models. After the approximation of the kernel tensor with the CP-decomposition, we calculate the accuracy of this decomposition, i.e. $\|K'-K\|/\|K\|$, where $K$ is the original tensor and $K'$ is the obtained approximation. The difference between the original tensor and our approximation may disturb data propagation in CNN, resulting in the drop of classification accuracy. We measure this drop before and after the fine-tuning of CNN. 

Furthermore, we record the CPU timings for our models and report the speed-up compared to the CPU timings of the original model (all timings are based on \texttt{Caffe} code run in the CPU mode on image batches of size 64). Finally, we report the reduction in the number of parameters resulting from the low-rank approximation. All results are reported for a number of ranks $R$.

\subsection{Character-classification CNN}

We use use CNN described in \citep{Jaderberg14a} for our experiments. The network has four convolutional layers with maxout nonlinearities between them and a softmax output. It was trained to classify $24\times24$ image patches into one of 36 classes (10 digits plus 26 characters). Our \texttt{Caffe} port of the publicly available pre-trained model (refered below as \texttt{CharNet}) achieves 91.2\% accuracy on test set (very similar to the original).

As in \citep{Jaderberg14}, we consider only the second and the third convolutional layers, which constitute more then 90\% of processing time. Layer 2 has 48 input and 128 output channels and filters of size $9\times9$, layer 3 has 64 input and 512 output channels, filter size is $8\times8$.

The results of separate approximation of layers 2 and 3 are shown in figures \ref{fig:janet_graph_2} and \ref{fig:janet_graph_3}. Tensor approximation error diminishes with the growth of approximation rank, and when the approximation rank becomes big enough, it is possible to approximate weight tensor accurately. However, our experiments showed that accurate approximation is not required for the network to function properly. For example, while approximating layer 3, network classification accuracy is unaffected even if tensor approximation error is as big as 78\%.

{\bf Combining approximations.}
We have applied our methods to two layers of the network using the following procedure. Firstly, layer 2 was approximated with rank 64. After that, the drop in accuracy was made small by fine-tuning of all layers but the new ones. Finally, layer 3 was approximated with rank 64, and for this layer such approximation does not result in significant drop of network prediction accuracy, so there is no need to fine-tune the network one more time. 

The network derived by this procedure is $8.5$ times faster than original model, while classification accuracy drops by $1\%$ to $90.2\%$.
Comparing with \citep{Jaderberg14}, we achieve almost two times bigger speedup for the same loss of accuracy, (\citep{Jaderberg14} incurs $1\%$ accuracy loss for the speedup of $4.2$x and $5\%$ accuracy loss for the speedup of $6$x).

\subsection{AlexNet}

\begin{figure}
  \centering
  \captionsetup[subfigure]{oneside,margin={1cm,0cm}}
  \begin{subfigure}[b]{0.38\textwidth}
    \centering
    \setlength\figureheight{14cm}
    \setlength\figurewidth{3.3cm}
%
%
\definecolor{mycolor1}{rgb}{0.00000,1.00000,1.00000}%
\definecolor{mycolor2}{rgb}{1.00000,0.00000,1.00000}%
\begin{tikzpicture}[font=\scriptsize]

\begin{axis}[%
width=\figurewidth,
height=0.23\figureheight,
at={(0\figurewidth,0.848075\figureheight)},
scale only axis,
xmin=4,
xmax=512,
xmode=log,
xtick={4, 16, 64, 256},
x tick label style={log ticks with fixed point},
ymin=0.2,
ymax=1.0,
ylabel={Approximation error},
myylabel absolute=-1cm
]
\addplot [color=blue,solid,mark=o,mark options={solid},forget plot,line width=1.0pt]
  table[row sep=crcr]{%
4       0.9718\\
8       0.9467\\
16      0.9031\\
32      0.8435\\
64      0.7687\\
128     0.6668\\
256     0.5214\\ 
512     0.3464\\
};
\end{axis}

\begin{loglogaxis}[%
log ticks with fixed point,
width=\figurewidth,
height=0.23\figureheight,
at={(0\figurewidth,0.565383\figureheight)},
scale only axis,
xmin=4,
xmax=512,
log basis x=2,
xtickten={2, 4, 6, 8},
ymin=0.01,
ymax=100,
extra y ticks={50},
extra y tick labels={50},
ytickten={-2, -1, 0, 1},
yminorticks=true,
ylabel={Accuracy drop (\%)},
myylabel absolute=-1cm,
legend style={at={($ (1,1) + (-0.1cm,-0.1cm) $)},anchor=north east,align=left,legend cell align=left,draw=black}
]
\addplot [color=red,dashed,mark=o,mark options={solid},line width=1.0pt]
  table[row sep=crcr]{%
4   83.5239\\
8  59.0255\\
16   18.9599\\
32    4.8588\\
64    1.9267\\
128    0.8079\\
256    0.3064\\
512   nan\\
};
\addlegendentry{no FT};

\addplot [color=red,solid,mark=o,mark options={solid},line width=1.0pt]
  table[row sep=crcr]{%
4    8.9098\\
8    5.5340\\
16    2.5054\\
32    0.6343\\
64    0.2292\\
128   0.01\\
256   0.01\\
512   0.01\\
};
\addlegendentry{FT};
\end{loglogaxis}

\begin{axis}[%
width=\figurewidth,
height=0.23\figureheight,
at={(0\figurewidth,0\figureheight)},
scale only axis,
xmin=4,
xmax=512,
xmode=log,
xtick={4, 16, 64, 256},
x tick label style={log ticks with fixed point},
xlabel={Rank},
ymin=1,
ymax=500,
ylabel={Parameters reduction ($ \times $)},
myylabel absolute=-1cm
]
\addplot [color=cyan,solid,mark=o,mark options={solid},forget plot,line width=1.0pt]
  table[row sep=crcr]{%
    4.0000  641.3196\\
    8.0000  320.6598\\
   16.0000  160.3299\\
   32.0000   80.1649\\
   64.0000   40.0825\\
  128.0000   20.0412\\
  256.0000   10.0206\\
  512.0000    5.0103\\
};
\end{axis}

\begin{axis}[%
width=\figurewidth,
height=0.23\figureheight,
at={(0\figurewidth,0.282692\figureheight)},
scale only axis,
xmin=4,
xmax=512,
xmode=log,
xtick={4, 16, 64, 256},
x tick label style={log ticks with fixed point},
ymin=1,
ymax=100,
ylabel={Speed-up ($ \times $)},
myylabel absolute=-1cm
]
\addplot [color=green!50!black,solid,mark=o,mark options={solid},forget plot,line width=1.0pt]
  table[row sep=crcr]{%
4   76.3863\\
8   35.1947\\
16   25.8771\\
32   15.7300\\
64    9.1373\\
128    4.8640\\
256    2.5607\\
512    1.2637\\
};
\end{axis}
\end{tikzpicture}%
    \caption{\texttt{CharNet conv2}}
    \label{fig:janet_graph_2}
  \end{subfigure}%
  \captionsetup[subfigure]{oneside,margin={0cm,0cm}}%
  \begin{subfigure}[b]{0.31\textwidth}
    \centering
    \setlength\figureheight{14cm}
    \setlength\figurewidth{3.3cm}
%
%
\definecolor{mycolor1}{rgb}{0.00000,1.00000,1.00000}%
\definecolor{mycolor2}{rgb}{1.00000,0.00000,1.00000}%
\begin{tikzpicture}[font=\scriptsize,trim axis left, trim axis right]

\begin{axis}[%
width=\figurewidth,
height=0.23\figureheight,
at={(0\figurewidth,0.848075\figureheight)},
scale only axis,
xmin=4,
xmax=512,
xmode=log,
xtick={4, 16, 64, 256},
x tick label style={log ticks with fixed point},
ymin=0.2,
ymax=1.0,
]
\addplot [color=blue,solid,mark=o,mark options={solid},forget plot,line width=1.0pt]
  table[row sep=crcr]{%
    4.0000    0.9651\\
    8.0000    0.9375\\
   16.0000    0.8961\\
   32.0000    0.8436\\
   64.0000    0.7827\\
  128.0000    0.7130\\
  256.0000    0.6284\\
  512.0000    0.5336\\
};
\end{axis}

\begin{loglogaxis}[%
log ticks with fixed point,
width=\figurewidth,
height=0.23\figureheight,
at={(0\figurewidth,0.565383\figureheight)},
scale only axis,
xmin=4,
xmax=512,
xtick={4, 16, 64, 256},
ymin=0.01,
ymax=100,
extra y ticks={50},
extra y tick labels={50},
ytickten={-2, -1, 0, 1},
yminorticks=true,
legend style={at={($ (1,1) + (-0.1cm,-0.1cm) $)},anchor=north east,align=left,legend cell align=left,draw=black}
]
\addplot [color=red,dashed,mark=o,mark options={solid},line width=1.0pt]
  table[row sep=crcr]{%
    4.0000   48.9753\\
    8.0000   19.9051\\
   16.0000    5.9584\\
   32.0000    1.6181\\
   64.0000    0.2678\\
  128.0000    0.0170\\
  256.0000    0.01\\
  512.0000    nan\\
};
\addlegendentry{no FT};

\addplot [color=red,solid,mark=o,mark options={solid},line width=1.0pt]
  table[row sep=crcr]{%
    4.0000    9.1605\\
    8.0000    3.2192\\
   16.0000    1.3094\\
   32.0000    0.8079\\
   64.0000    0.01\\
  128.0000    0.01\\
  256.0000    0.01\\
  512.0000    0.01\\
};
\addlegendentry{FT};
\end{loglogaxis}

\begin{axis}[%
width=\figurewidth,
height=0.23\figureheight,
at={(0\figurewidth,0\figureheight)},
scale only axis,
xmin=4,
xmax=512,
xmode=log,
xtick={4, 16, 64, 256},
x tick label style={log ticks with fixed point},
xlabel={Rank},
ymin=1,
ymax=500,
]
\addplot [color=cyan,solid,mark=o,mark options={solid},forget plot,line width=1.0pt]
  table[row sep=crcr]{%
    4.0000  630.1538\\
    8.0000  315.0769\\
   16.0000  157.5385\\
   32.0000   78.7692\\
   64.0000   39.3846\\
  128.0000   19.6923\\
  256.0000    9.8462\\
  512.0000    4.9231\\
};
\end{axis}

\begin{axis}[%
width=\figurewidth,
height=0.23\figureheight,
at={(0\figurewidth,0.282692\figureheight)},
scale only axis,
xmin=4,
xmax=512,
xmode=log,
xtick={4, 16, 64, 256},
x tick label style={log ticks with fixed point},
ymin=1,
ymax=100,
]
\addplot [color=green!50!black,solid,mark=o,mark options={solid},forget plot,line width=1.0pt]
  table[row sep=crcr]{%
    4.0000  410.2302\\
    8.0000  258.2471\\
   16.0000  141.4924\\
   32.0000   77.4827\\
   64.0000   60.4232\\
  128.0000   29.9510\\
  256.0000   12.4727\\
  512.0000    5.8099\\
};
\end{axis}
\end{tikzpicture}%
    \caption{\texttt{CharNet conv3}}
    \label{fig:janet_graph_3}
  \end{subfigure}%
  \captionsetup[subfigure]{oneside,margin={0cm,0cm}}%
  \begin{subfigure}[b]{0.31\textwidth}
    \centering
    \setlength\figureheight{14cm}
    \setlength\figurewidth{3.3cm}
%
%
\definecolor{mycolor1}{rgb}{0.00000,1.00000,1.00000}%
\definecolor{mycolor2}{rgb}{1.00000,0.00000,1.00000}%
\begin{tikzpicture}[font=\scriptsize,trim axis left, trim axis right]

\begin{axis}[%
width=\figurewidth,
height=0.23\figureheight,
at={(0\figurewidth,0.848075\figureheight)},
scale only axis,
xmin=100,
xmax=300,
xtick={100, 140, 200, 250, 300},
ymin=0.2,
ymax=1,
]
\addplot [color=blue,solid,mark=o,mark options={solid},forget plot,line width=1.0pt]
  table[row sep=crcr]{%
100	0.8037\\
140	0.5035\\
200	0.4234\\
250	0.3734\\
300	0.3336\\
};
\end{axis}

\begin{axis}[%
log ticks with fixed point,
width=\figurewidth,
height=0.23\figureheight,
at={(0\figurewidth,0.565383\figureheight)},
scale only axis,
xmin=100,
xmax=300,
xtick={100, 140, 200, 250, 300},
ymode=log,
ymin=0,
ymax=100,
yminorticks=true,
legend style={at={($ (1,1) + (-0.1cm,-0.1cm) $)},anchor=north east,align=left,legend cell align=left,draw=black}
]
\addplot [color=red,dashed,mark=o,mark options={solid},line width=1.0pt]
  table[row sep=crcr]{%
100	69.202\\
140	3.204\\
200	0.970000000000004\\
250	0.580000000000003\\
300	0.3\\
};
\addlegendentry{no FT};

\addplot [color=red,solid,mark=o,mark options={solid},line width=1.0pt]
  table[row sep=crcr]{%
100	18.95\\
140	1.23\\
200	0.449999999999995\\
250	nan\\
300	nan\\
};
\addlegendentry{FT};
\end{axis}

\begin{axis}[%
width=\figurewidth,
height=0.23\figureheight,
at={(0\figurewidth,0\figureheight)},
scale only axis,
xlabel={Rank},
xmin=100,
xmax=300,
xtick={100, 140, 200, 250, 300},
ymin=4,
ymax=20,
]
\addplot [color=cyan,solid,mark=o,mark options={solid},forget plot,line width=1.0pt]
  table[row sep=crcr]{%
100	16.9724\\
140	12.1231\\
200	8.4862\\
250	6.7890\\
300	5.6575\\
};
\end{axis}

\begin{axis}[%
width=\figurewidth,
height=0.23\figureheight,
at={(0\figurewidth,0.282692\figureheight)},
scale only axis,
xmin=100,
xmax=300,
xtick={100, 140, 200, 250, 300},
ymin=1,
ymax=7,
]
\addplot [color=green!50!black,solid,mark=o,mark options={solid},forget plot,line width=1.0pt]
  table[row sep=crcr]{%
100	6.51619952494062\\
140	4.52693069306931\\
200	3.20856140350877\\
250	1.75292012779553\\
300	1.564928693668\\
};
\end{axis}
\end{tikzpicture}%
    \caption{\texttt{AlexNet conv2}}
    \label{fig:alexnet_graph}
  \end{subfigure}
  \caption{{\bf Various properties and performance metrics of different approximated CNNs plotted as functions of the approximation rank.} {\bf First row:} kernel tensor approximation error. {\bf Second row:} drop of the classification accuracy of the full model with approximated layers w.r.t the accuracy of the original model; {\it dashed} lines correspond to the non-tuned CNNs, {\it solid} lines depict the performance after the fine-tuning. \textit{Note the log-scale. Cases, where the accuracy has actually improved are plotted at the bottom line.} {\bf Third row:} empirical speed-up of the approximated layer w.r.t. the original layer. {\bf Fourth row:} ratio between the numbers of parameters in the original and the approximated layers.}
  \label{fig:graphs}
\end{figure}
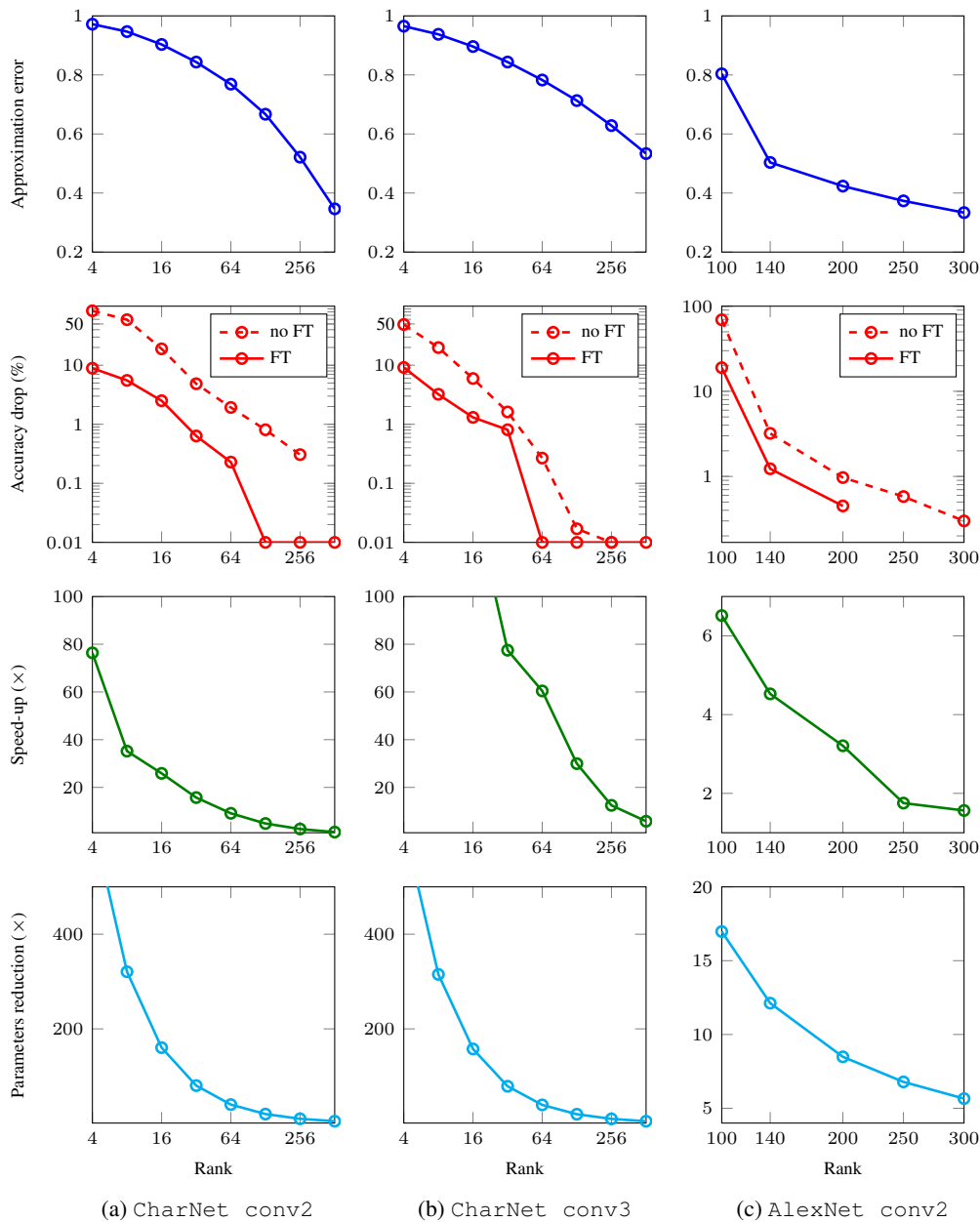

Following \cite{Denton14} we also consider the second convolutional layer of \texttt{AlexNet} \cite{Krizhevsky12}. As a baseline we use a pre-trained model shipped with \texttt{Caffe}. We summarize various network properties for several different ranks of approximation in \fig{alexnet_graph}. It can be noticed that \texttt{conv2} of the considered network demands far larger rank (comparing to the \texttt{CharNet} experiment) for achieving proper performance. 

Overall, in order to reach the $ 0.5\% $ accuracy drop reported in \citep{Jaderberg14a} it is sufficient to take $200$ components, which also gives a superior layer speed-up ($ 3.6\times $ vs. $ 2\times $ achieved by Scheme 2 of \citep{Jaderberg14a}). The running time of the \texttt{conv2} can be further reduced if we allow for slightly more misclassifications: rank 140 approximation leads to $ 4.5\times $ speed-up at the cost of $ \approx 1\% $ accuracy loss surpassing the results of \citep{Denton14}.

Along with conventional full-network fine-tuning we tried to refine the obtained tensor approximation by applying the data reconstruction approach from \citep{Jaderberg14a}. Unfortunately, we failed to find a good SGD learning rate: larger values led to the exploding gradients, while the smaller ones did not allow to sensibly reduce the reconstruction loss. We suspect that this effect is due to the instability of the low-rank CP-decomposition \citep{De08}. One way to circumvent the issue would be to alternate the components learning (i.e. not optimizing all of them at once), which is the scope of our future work.

Our latest experiments showed that while our approach is superior for smaller architectures (as in character classification), it is not the best one for large nets such as AlexNet. Although \citep{Jaderberg14a} mentions that application of their approach to the second layer of the OverFeat architecture yields a $ 2\times $ speed-up, our colleagues at Yandex have discovered that a far greater speed-up with \citep{Jaderberg14a} can be reached, at least for AlexNet. In particular, our experiments with \citep{Jaderberg14a} on AlexNet have showed that the second convolutional layer of AlexNet can be accelerated by the factor of $ 6.6\times$ at the cost of $ \approx 1\% $ accuracy loss via Scheme 2 and data optimization as described in \citep{Jaderberg14a} (as compared to $4.5\times$ for similar accuracy loss obtainable with our method).

\subsection{NLS vs. Greedy}

One of our main contributions is pointing out that greedy CP-decomposition works worse than more advanced algorithms such as non-linear least squares (NLS), and evaluate this degradation in the context of speeding up CNNs.

We perform comparisons on the second layers of both \texttt{CharNet} and \texttt{AlexNet}. For \texttt{CharNet}, we also evaluate the combination with fine-tuning. Furthermore, for \texttt{CharNet}, we also tried initializing the layers randomly (using the scheme of \citep{Glorot10}).

The results in \tab{greedy} clearly demonstrate two related things. Firstly, NLS decomposition leads to significantly higher accuracy whether with fine-tuning or without. The advantage is greater for the more complex network (\texttt{AlexNet}). 

Secondly, the output of the fine-tuning clearly depends on the quality of the approximation. This observation concurs with the hypothesis that the large number of parameters in CNN is needed to avoid poor local minima. Indeed, CP-decomposition radically decreases the number of parameters in a layer. While good minima may still exist (e.g.\ the NLS+FT result), optimization is prone to stucking in a much worse minima (e.g.\ the Random+FT result).



%
%
%

\begin{table}[h!]
\setlength{\tabcolsep}{3pt}
  \caption{{\bf Accuracy drop for the greedy and the non-linear least-squares (NLS) CP-decomposition.} The results are given for the second layers of \texttt{CharNet} and \texttt{AlexNet}, for different decomposition ranks $R$, and the numbers correspond to the accuracy drop of the entire CNN. Original networks achieve $91.24\%$ (\texttt{CharNet}) and $79.95\%$ (\texttt{AlexNet}). For \texttt{CharNet} we also evaluate the effect of fine-tuning (FT), and also for the random initialization. NLS computation of the CP-decomposition invariably leads to better performance, and the advantage persists through fine-tuning.  }
  \label{tab:greedy}
  \begin{center}
  \begin{sc}
    \renewcommand{\arraystretch}{1.3}
    \begin{tabular}{c | c c | c c | c c| c | c | c }
             & \multicolumn{2}{c |}{\small\texttt{\textup{CharNet}}, R=16} &
               \multicolumn{2}{c |}{\small\texttt{\textup{CharNet}}, R=64} &
               \multicolumn{2}{c |}{\small\texttt{\textup{CharNet}}, R=256} &
               \small\texttt{\textup{AlexNet}} &         
               \small\texttt{\textup{AlexNet}} &
               \small\texttt{\textup{AlexNet}} \\
      \cline{2-7} 
             & \small no FT & \small FT &
               \small no FT & \small FT &
               \small no FT &  \small FT & \small R=140 & \small R=200 & \small R=300 \\
      \hline
       Random & --  & 9.70  & 
                --  & 7.64 & 
                --  & 6.13 & -- & -- & -- \\
       Greedy & 24.15 & 2.64 & 
                 4.99 & 0.46 & 
                 1.14 & -0.31 & 65.04 & 7.29 & 4.76 \\
       NLS    & \bf 18.96 & \bf  2.16 & 
               \bf  1.93 & \bf  0.09 & 
               \bf  0.31 & \bf -0.52 & \bf 3.21 & \bf 0.97 & \bf 0.30\\
    \end{tabular}
  \end{sc}
  \end{center}
\end{table}

\section{Discussion}
\label{sect:discussion}

We have demonstrated that a rather straightforward application of a modern tensor decomposition method (NLS-based low-rank CP-decomposition) combined with the discriminative fine-tuning of the entire network can achieve considerable speedups with minimal loss in accuracy. 

In the preliminary comparisons, this approach outperforms the previous methods of \citep{Denton14} and \citep{Jaderberg14}, for the character classification CNN. However, more comparisons especially with a more related work of \citep{Denton14} are needed. In particular, it is to be determined whether bi-clustering is useful, when non-linear least squares are used for CP-decomposition.

Another avenue of research are layers with spatially-varying kernels, such as used e.g.\ in \citep{Taigman14}. Firstly, these layers would greatly benefit from the reduction in the number of parameters. Secondly, the spatial variation of the kernel might be embedded into extra tensor dimensions, which may open up further speed-up possibilities.

Finally, similarly to \citep{Denton14}, we note that low-rank decompositions seems to have a regularizing effects allowing to slightly improve the overall accuracy for higher rank. 

\section*{Acknowledgments}
The work presented in Section 3 was supported by Russian Science Foundation grant 14-11-00659.



\bibliography{iclr2015}

\begin{thebibliography}{22}
\providecommand{\natexlab}[1]{#1}
\providecommand{\url}[1]{\texttt{#1}}
\expandafter\ifx\csname urlstyle\endcsname\relax
  \providecommand{\doi}[1]{doi: #1}\else
  \providecommand{\doi}{doi: \begingroup \urlstyle{rm}\Url}\fi

\bibitem[Chellapilla et~al.(2006)Chellapilla, Puri, and Simard]{Chellapilla06}
Chellapilla, Kumar, Puri, Sidd, and Simard, Patrice.
\newblock High performance convolutional neural networks for document
  processing.
\newblock In \emph{Tenth International Workshop on Frontiers in Handwriting
  Recognition}, 2006.

\bibitem[Chintala(2014)]{Chintala14}
Chintala, Soumith.
\newblock Convnet-benchmarks.
\newblock \url{https://github.com/soumith/convnet-benchmarks}, 2014.
\newblock Accessed: 2014-12-19.

\bibitem[De~Silva \& Lim(2008)De~Silva and Lim]{De08}
De~Silva, Vin and Lim, Lek-Heng.
\newblock Tensor rank and the ill-posedness of the best low-rank approximation
  problem.
\newblock \emph{SIAM J. Matrix Anal. Appl.}, 30\penalty0 (3):\penalty0
  1084--1127, 2008.

\bibitem[Dean et~al.(2012)Dean, Corrado, Monga, Chen, Devin, Mao, Senior,
  Tucker, Yang, Le, et~al.]{Dean12}
Dean, Jeffrey, Corrado, Greg, Monga, Rajat, Chen, Kai, Devin, Matthieu, Mao,
  Mark, Senior, Andrew, Tucker, Paul, Yang, Ke, Le, Quoc~V, et~al.
\newblock Large scale distributed deep networks.
\newblock In \emph{Advances in Neural Information Processing Systems}, pp.\
  1223--1231, 2012.

\bibitem[Denton et~al.(2014)Denton, Zaremba, Bruna, LeCun, and
  Fergus]{Denton14}
Denton, Emily, Zaremba, Wojciech, Bruna, Joan, LeCun, Yann, and Fergus, Rob.
\newblock Exploiting linear structure within convolutional networks for
  efficient evaluation.
\newblock \emph{arXiv preprint arXiv:1404.0736}, 2014.

\bibitem[Farabet et~al.(2011)Farabet, LeCun, Kavukcuoglu, Culurciello, Martini,
  Akselrod, and Talay]{Farabet11}
Farabet, Cl{\'e}ment, LeCun, Yann, Kavukcuoglu, Koray, Culurciello, Eugenio,
  Martini, Berin, Akselrod, Polina, and Talay, Selcuk.
\newblock Large-scale {FPGA}-based convolutional networks.
\newblock \emph{Machine Learning on Very Large Data Sets}, 2011.

\bibitem[Glorot \& Bengio(2010)Glorot and Bengio]{Glorot10}
Glorot, Xavier and Bengio, Yoshua.
\newblock Understanding the difficulty of training deep feedforward neural
  networks.
\newblock In \emph{Proceedings of the Thirteenth International Conference on
  Artificial Intelligence and Statistics, {AISTATS} 2010, Chia Laguna Resort,
  Sardinia, Italy, May 13-15, 2010}, pp.\  249--256, 2010.

\bibitem[Jaderberg et~al.(2014{\natexlab{a}})Jaderberg, Vedaldi, and
  Zisserman]{Jaderberg14}
Jaderberg, Max, Vedaldi, Andrea, and Zisserman, Andrew.
\newblock Speeding up convolutional neural networks with low rank expansions.
\newblock In \emph{Proceedings of the British Machine Vision Conference
  (BMVC)}, 2014{\natexlab{a}}.

\bibitem[Jaderberg et~al.(2014{\natexlab{b}})Jaderberg, Vedaldi, and
  Zisserman]{Jaderberg14a}
Jaderberg, Max, Vedaldi, Andrea, and Zisserman, Andrew.
\newblock Deep features for text spotting.
\newblock In \emph{Computer Vision--ECCV 2014}, pp.\  512--528. Springer,
  2014{\natexlab{b}}.

\bibitem[Jia et~al.(2014)Jia, Shelhamer, Donahue, Karayev, Long, Girshick,
  Guadarrama, and Darrell]{caffe}
Jia, Yangqing, Shelhamer, Evan, Donahue, Jeff, Karayev, Sergey, Long, Jonathan,
  Girshick, Ross, Guadarrama, Sergio, and Darrell, Trevor.
\newblock Caffe: Convolutional architecture for fast feature embedding.
\newblock \emph{arXiv preprint arXiv:1408.5093}, 2014.

\bibitem[Kofidis \& Regalia(2002)Kofidis and Regalia]{Kof02}
Kofidis, Eleftherios and Regalia, Phillip~A.
\newblock On the best rank-1 approximation of higher-order supersymmetric
  tensors.
\newblock \emph{SIAM J. Matrix Anal. Appl.}, 23\penalty0 (3):\penalty0
  863--884, 2002.

\bibitem[Kolda \& Bader(2009)Kolda and Bader]{kolda-review-2009}
Kolda, T.~G. and Bader, B.~W.
\newblock Tensor decompositions and applications.
\newblock \emph{SIAM Rev.}, 51\penalty0 (3):\penalty0 455--500, 2009.

\bibitem[Krizhevsky et~al.(2012)Krizhevsky, Sutskever, and
  Hinton]{Krizhevsky12}
Krizhevsky, Alex, Sutskever, Ilya, and Hinton, Geoffrey~E.
\newblock Imagenet classification with deep convolutional neural networks.
\newblock In \emph{Advances in neural information processing systems}, pp.\
  1097--1105, 2012.

\bibitem[LeCun et~al.(1989)LeCun, Boser, Denker, Henderson, Howard, Hubbard,
  and Jackel]{LeCun89}
LeCun, Yann, Boser, Bernhard, Denker, John~S, Henderson, Donnie, Howard,
  Richard~E, Hubbard, Wayne, and Jackel, Lawrence~D.
\newblock Backpropagation applied to handwritten zip code recognition.
\newblock \emph{Neural computation}, 1\penalty0 (4):\penalty0 541--551, 1989.

\bibitem[Lin et~al.(2013)Lin, Chen, and Yan]{Lin13}
Lin, Min, Chen, Qiang, and Yan, Shuicheng.
\newblock Network in network.
\newblock \emph{arXiv preprint arXiv:1312.4400}, 2013.

\bibitem[Mathieu et~al.(2013)Mathieu, Henaff, and LeCun]{Mathieu13}
Mathieu, Michael, Henaff, Mikael, and LeCun, Yann.
\newblock Fast training of convolutional networks through {FFT}s.
\newblock \emph{arXiv preprint arXiv:1312.5851}, 2013.

\bibitem[Rigamonti et~al.(2013)Rigamonti, Sironi, Lepetit, and
  Fua]{Rigamonti13}
Rigamonti, Roberto, Sironi, Amos, Lepetit, Vincent, and Fua, Pascal.
\newblock Learning separable filters.
\newblock In \emph{Computer Vision and Pattern Recognition (CVPR), 2013 IEEE
  Conference on}, pp.\  2754--2761. Ieee, 2013.

\bibitem[Simonyan \& Zisserman(2014)Simonyan and Zisserman]{Simonyan14}
Simonyan, Karen and Zisserman, Andrew.
\newblock Very deep convolutional networks for large-scale image recognition.
\newblock \emph{CoRR}, abs/1409.1556, 2014.
\newblock URL \url{http://arxiv.org/abs/1409.1556}.

\bibitem[Sorber et~al.(2014)Sorber, Van~Barel, and
  De~Lathauwer]{sorber-tensorlab-2014}
Sorber, L., Van~Barel, M., and De~Lathauwer, L.
\newblock Tensorlab v2.0.
\newblock \emph{Available online}, 2014.
\newblock URL \url{http://tensorlab.net}.

\bibitem[Stegeman \& Comon(2010)Stegeman and Comon]{Stegeman10}
Stegeman, Alwin and Comon, Pierre.
\newblock Subtracting a best rank-1 approximation may increase tensor rank.
\newblock \emph{Linear Algebra Appl.}, 433\penalty0 (7):\penalty0 1276--1300,
  2010.

\bibitem[Taigman et~al.(2014)Taigman, Yang, Ranzato, and Wolf]{Taigman14}
Taigman, Yaniv, Yang, Ming, Ranzato, Marc'Aurelio, and Wolf, Lior.
\newblock Deepface: Closing the gap to human-level performance in face
  verification.
\newblock In \emph{Computer Vision and Pattern Recognition (CVPR), 2014 IEEE
  Conference on}, pp.\  1701--1708. IEEE, 2014.

\bibitem[Tomasi \& Bro(2006)Tomasi and Bro]{tomasi-comparison-2006}
Tomasi, Giorgio and Bro, Rasmus.
\newblock A comparison of algorithms for fitting the parafac model.
\newblock \emph{Comp. Stat. Data An.}, 50\penalty0 (7):\penalty0 1700--1734,
  2006.

\end{thebibliography}
\bibliographystyle{iclr2015}

\pagebreak
\section*{Appendix}

In all our experiments, we consistently found that  non-linear least squares (NLS) with the implementation from \citep{sorber-tensorlab-2014} yield better CP-decompositions with smaller ranks (for the same approximation accuracy) than the greedy approach of finding the next best rank-one tensor and adding it to the decomposition.

Such advantage can be highlighted by the following example of $2\times 2 \times 2$ tensor $G$ of rank two, whose frontal slices are defined by:
$$
G_1 = 
    \begin{pmatrix}
     1 & 0 \\
     0 & 1
    \end{pmatrix},
\quad 
G_2 = 
    \begin{pmatrix}
     1 & 1 \\
     0 & 2
    \end{pmatrix}.
$$
We checked numerically that a sequence of two best rank-one approximations fails to approximate the tensor $G$ -- relative error was $0.35$. In contrast to that, the best rank-two approximation (with $10^{-7}$ error in this case) was successfully obtained by the NLS optimization.

\end{document}